\documentclass[letterpaper, 10 pt, journal, twoside]{ieeetran}
\usepackage{amsmath,amsfonts}
\usepackage{algorithmic}
\usepackage{array}
\usepackage[caption=false,font=normalsize,labelfont=sf,textfont=sf]{subfig}
\usepackage{textcomp}
\usepackage{stfloats}
\usepackage{url}
\usepackage{verbatim}
\usepackage{graphicx}
\usepackage[ruled,vlined]{algorithm2e}
\usepackage{multirow}
\usepackage{cite}
\usepackage{placeins} 
\usepackage{hyperref}
\usepackage{orcidlink}
\usepackage{comment}
\usepackage{makecell}
\SetKw{KwReturn}{return}
\begin{document}

\bstctlcite{IEEEexample:BSTcontrol}

\title{ModuLoop : Low-Level Code Generation using Modular Synthesizer and Closed-Loop Debugger for Robotic Control}

\author{
  Gina Yoon\,\orcidlink{0009-0009-2287-9945},
  \IEEEmembership{Student Member, IEEE}, 
  \and
  Sumin Lee\,\orcidlink{0009-0005-9994-8444},
  \IEEEmembership{Student Member, IEEE}, 
  \and
  Joo Yong Sim\,\orcidlink{0000-0003-3779-7589},
  \IEEEmembership{Member, IEEE}
  \thanks{Manuscript received: May, 24, 2025; Revised: August, 29, 2025; Accepted; October, 3, 2025.}
  \thanks{This paper was recommended for publication by Editor Chao-Bo Yan upon evaluation of the Associate Editor and Reviewers’ comments.}
  \thanks{This work was supported by  National Research Foundation of Korea(NRF) grant (No. RS-2025-02216282, RS-2025-16070288), Institute of Information \& Communications Technology Planning \& Evaluation (IITP) grant funded by the Korea government (MSIT) (No. RS-2022-II220025) and by Ministry of Trade, Industry and Energy (MOTIE RS 2023 00258591). \textit{(Gina Yoon and Sumin Lee are co-first authors.) (Corresponding author: Joo Yong Sim.)}}
  \thanks{Gina Yoon, Sumin Lee, and Joo Yong Sim are with the Department of Mechanical Systems Engineering, Sookmyung Women's University, Cheongpa-ro 47-gil 100, Yongsan-gu, Seoul, 04310, South Korea (e-mail: ppippi272@sookmyung.ac.kr, sstone@sookmyung.ac.kr, jysim@sookmyung.ac.kr).}

     \thanks{Digital Object Identifier (DOI): see top of this page.}
 }

%\markboth{IEEE ROBOTICS AND AUTOMATION LETTERS. PREPRINT VERSION. 05, 2025}%
%\markboth{IEEE Robotics and Automation Letters. Preprint Version. Submitted for review, May 2025}{}%

\maketitle

\begin{abstract}

Large Language Models (LLMs) have demonstrated impressive performance across various domains, including code generation and problem solving. However, their application in robotic control—particularly in low-level tasks that require precise manipulation, real-time feedback, and environment-dependent execution—remains limited.
To address this challenge, we propose the Closed-Loop Modular Code Synthesizer framework. This framework leverages a pre-trained LLM without any task-specific fine-tuning to perform modular code planning and generation, and iteratively executes the generated code while inserting debugging probes to observe its behavior. This closed-loop structure facilitates systematic debugging and refinement, ultimately producing executable control programs.
We apply the proposed framework to the calibration of an RGB-D camera and a robotic arm, validating its effectiveness in real-world settings. Furthermore, through a subsequent pick-and-place task, we demonstrate not only the accuracy of the calibration but also the potential extensibility of the framework. Across both tasks, the framework achieved high execution accuracy and autonomy, illustrating the practicality and scalability of LLM-based robotic control using our framework.

\end{abstract}

\section{Introduction}

\IEEEPARstart{L}{arge} Language Models (LLMs) have demonstrated remarkable capabilities across a range of domains, including natural language understanding, code generation, and symbolic reasoning. Advanced models such as the GPT series~\cite{openai2024gpt4o} have shown the ability to generate executable code from natural language instructions, highlighting their growing utility in industrial and applied settings~\cite{jiang2024survey}.  In the field of
 robotics, recent approaches have sought to enhance LLMs’
 task interpretation and planning capabilities through large scale dataset fine-tuning or few-shot learning\cite{brown2020language, pmlr-v229-zitkovich23a, liang2023code}.

Against this backdrop, recent LLM-based robotic control studies have generally progressed in two directions: (i) translating natural language commands into high-level plans while relying on external modules for execution~\cite{ahn2022can, pmlr-v229-zitkovich23a}, and (ii) directly generating executable low-level control or perception code through user-defined APIs from natural language instructions~\cite{liang2023code, singh2023progprompt, huang2023voxposer}. While the former limits LLM involvement in execution, the latter often suffers from prompt dependency and poor generalization—issues particularly critical in low-level control where precise motion and real-time error handling are required.

\begin{figure*}
    \centering
    \includegraphics[width=1.0\linewidth]{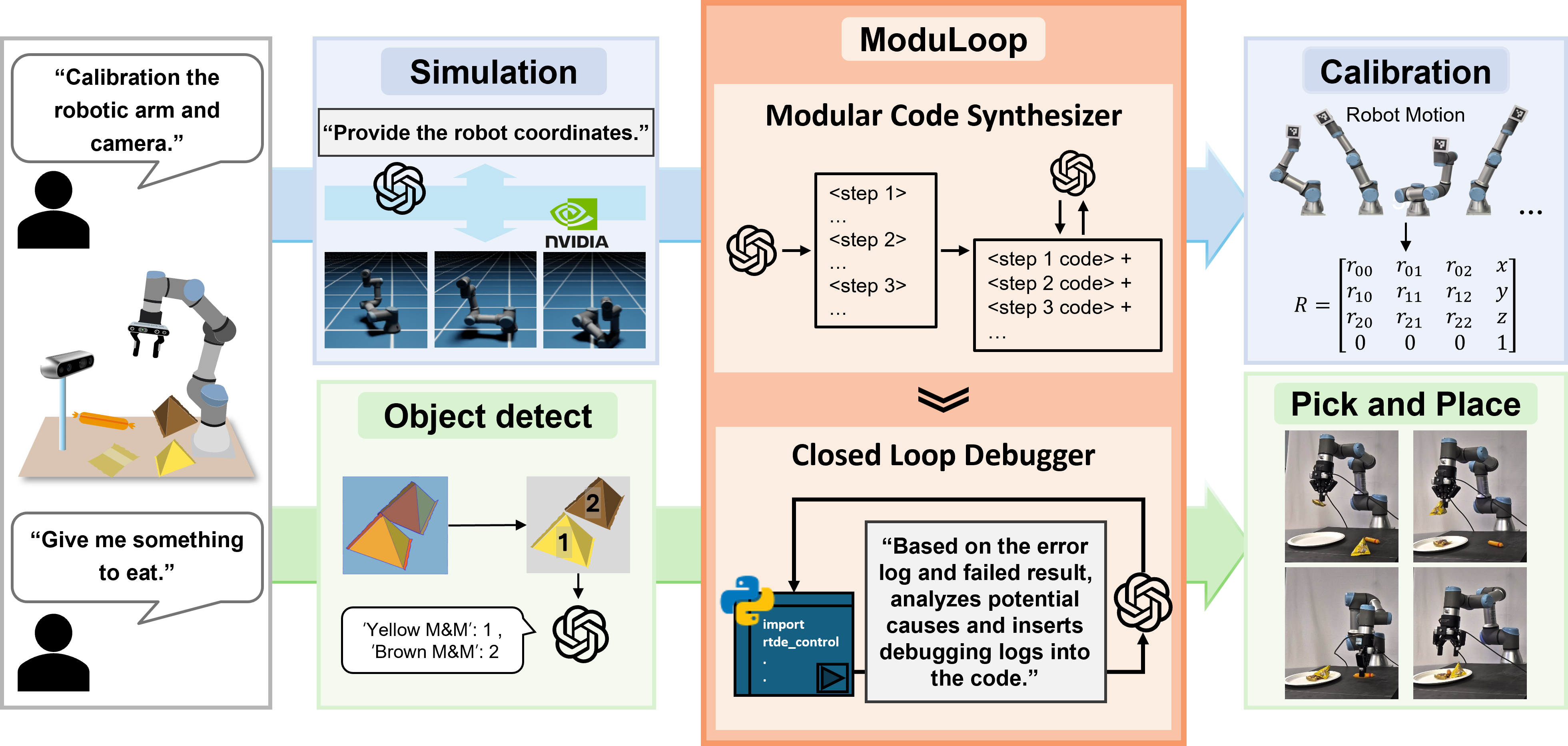}
    \caption{The architecture of an LLM-based control framework that enables a robotic arm to autonomously execute tasks based on natural language commands. ModuLoop serves as a code synthesis framework that generates executable control code for both low-level hand–eye calibration and high-level pick-and-place manipulation. Before code generation, the system collects task-specific environmental information — simulation-based reachable coordinate verification for calibration, and SAM- and LLM-based object detection for pick-and-place. Using these inputs, ModuLoop performs modular code synthesis and closed-loop debugging, producing verified control code that enables accurate calibration and autonomous manipulation.}
    \label{fig:over-view}
\end{figure*}

To address these challenges and enable LLMs to participate actively in the control loop, we propose \textbf{ModuLoop}, a framework that enables LLMs to participate in the full control loop of low-level robotic tasks. Fig.~\ref{fig:over-view} illustrates the overall architecture of the ModuLoop based robotic control system. The pipeline begins with a high-level task command expressed in natural language. ModuLoop decomposes the given instruction into fine-grained subtasks and converts each subtask into executable Python code.
Based on execution feedback—including runtime errors and performance metrics—the framework dynamically refines and corrects the generated code, forming a closed-loop structure in which the robot autonomously learns and improves through iterative execution.

We validate ModuLoop in two representative tasks: (1)
hand–eye calibration between a camera and robotic manipulator, and (2) a pick-and-place manipulation task requiring perception and motion planning. Although originally designed for calibration without any user-defined APIs, the framework demonstrated extensibility to manipulation tasks with minimal perception guidance.

Our main contributions are threefold:
\begin{enumerate}
    \item A simulation-based LLM feedback loop that secures collision-free executable coordinates for hand–eye calibration, thereby enhancing data efficiency and accuracy.
    \item A modular code synthesizer framework that decomposes high-level natural language commands into executable low-level Python modules for robot control.
    \item A closed-loop debugging mechanism that uses execution feedback—such as runtime errors and accuracy metrics—to autonomously revise and improve generated code.

\end{enumerate}

Together, these contributions position ModuLoop as a prac-
tical and generalizable framework that elevates LLMs from
passive planners to autonomous agents capable of generating, executing, and adapting robot control code in real-world environments.

% \begin{figure*}
%     \centering
%     \includegraphics[width=1.0\linewidth]{figures/Figure1.png}
%     \caption{The architecture of an LLM-based control framework that enables a robotic arm to autonomously execute tasks based on natural language commands. The system comprises four stages: (1) LLM-based robot coordinate generation and simulation verification. (2) Code generation and execution for hand–eye calibration via ModuLoop. (3) LLM-based object recognition. (4) Code generation and execution for pick-and-place tasks via ModuLoop. Upon receiving a user command as a natural language prompt, the LLM and NVIDIA Isaac Sim iteratively generate and verify feasible coordinates. The validated coordinates are then passed to ModuLoop, which proceeds to the code generation stage. ModuLoop plans and refines the code through modular synthesis and closed-loop debugging, computing the transformation matrix in the process. The SAM and LLM-based recognition modules subsequently extract the depth and rotation of the target object. The object information and transformation matrix are returned to ModuLoop, which converts the initial user command into executable control code. Finally, the robot grasps the target object and moves it to the designated location, completing the task.}
%     \label{fig:over-view}
% \end{figure*}

\section{Related Work}

\subsection{LLM-Driven Robotic Control}

Recent research on integrating Large Language Models (LLMs) into robotic systems has generally followed two directions. The first focuses on high-level action planning, as seen in systems such as \textit{SayCan}~\cite{ahn2022can} and \textit{PaLM-E, RT-1, RT-2}~\cite{driess2023palm, brohan2022rt, pmlr-v229-zitkovich23a}, which combine LLMs with affordance models or multimodal inputs to translate language commands into abstract action sequences. However, these approaches do not produce executable low-level control code and instead rely on downstream modules.

The second direction explores directly generating robot control code from natural language. \textit{Code-as-Policies}~\cite{liang2023code}, for example, uses few-shot prompting with API specifications and annotated examples, while \textit{ProgPrompt}~\cite{singh2023progprompt} employs structured prompts with action primitives, and \textit{RobotGPT}~\cite{jin2024robotgpt} generates and simulates code via ChatGPT with reinforcement learning. Despite producing executable code, most approaches remain limited by their reliance on structured inputs and lack of feedback-based refinement, reducing adaptability across environments. RobotScript~\cite{robotscript} also demonstrates LLM-based code generation. It not only integrates perception tools like AnyGrasp~\cite{anygrasp} for grip pose prediction but also exposes them at the API level for direct invocation within the generated script.

\subsection{Closed-Loop Refinement for Code Generation using LLM}

Existing LLM-based code generation systems (e.g., Codex~\cite{chen2021evaluating}) operate in an open-loop manner, producing static code without execution feedback. \textit{AutoGen} \cite{autogen} introduces a multi-agent framework, while \textit{Reflexion}\cite{reflexion} allows a single LLM to refine reasoning through self-reflection, both enabling iterative improvement. However, these methods remain confined to abstract programming and lack applicability to robotic control, where continuous sensing and physical feedback are essential. \textit{MCCoder}~\cite{mccoder} extends this line by incorporating sensor feedback for microcontroller-based robots, but its reliance on hardware-specific prompts and rigid templates limits generalizability. In contrast, ModuLoop autonomously synthesizes and improves executable control code by directly leveraging real-world feedback.

\subsection{Hand-eye Calibration}
Hand-eye calibration~\cite{horaud1995hand} estimates the spatial transformation between a robot’s end-effector and a mounted camera, enabling vision-based manipulation. It aligns the coordinate frames of the robot and the camera so that perceived object positions can be accurately used for motion control. Recent learning-based methods automate this process, either by regressing extrinsics from RGB or point clouds without markers~\cite{falisse2025marker}, or by aligning point clouds via registration pipelines~\cite{tang2024kalib, li2024automatic}. These reduce marker dependence and manual effort but remain sensitive to training data distribution and  rely on accurate initial alignment for reliable performance. ModuLoop overcomes these issues by enabling automatic, adaptive calibration without prior data or manual tuning.

\section{ Low-Level Control with Workspace Validation via Simulation and Closed-Loop Modular Code Generation}
\subsection{Iterative Interaction between LLM and Simulator for Physically Valid Motion Planning}

To verify the physical feasibility of motion trajectories prior to real-world execution, we employ a simulation-based validation environment in Isaac Sim~\cite{isaacsim} to check workspace reachability and potential collisions. As shown in Fig.~\ref{fig:simul_gpt}, within each iteration of the LLM–simulator feedback loop, the LLM generates calibration targets—specifically, a set of 3D end-effector position candidates:

\begin{figure}
    \centering
    \includegraphics[width=1.0\linewidth]{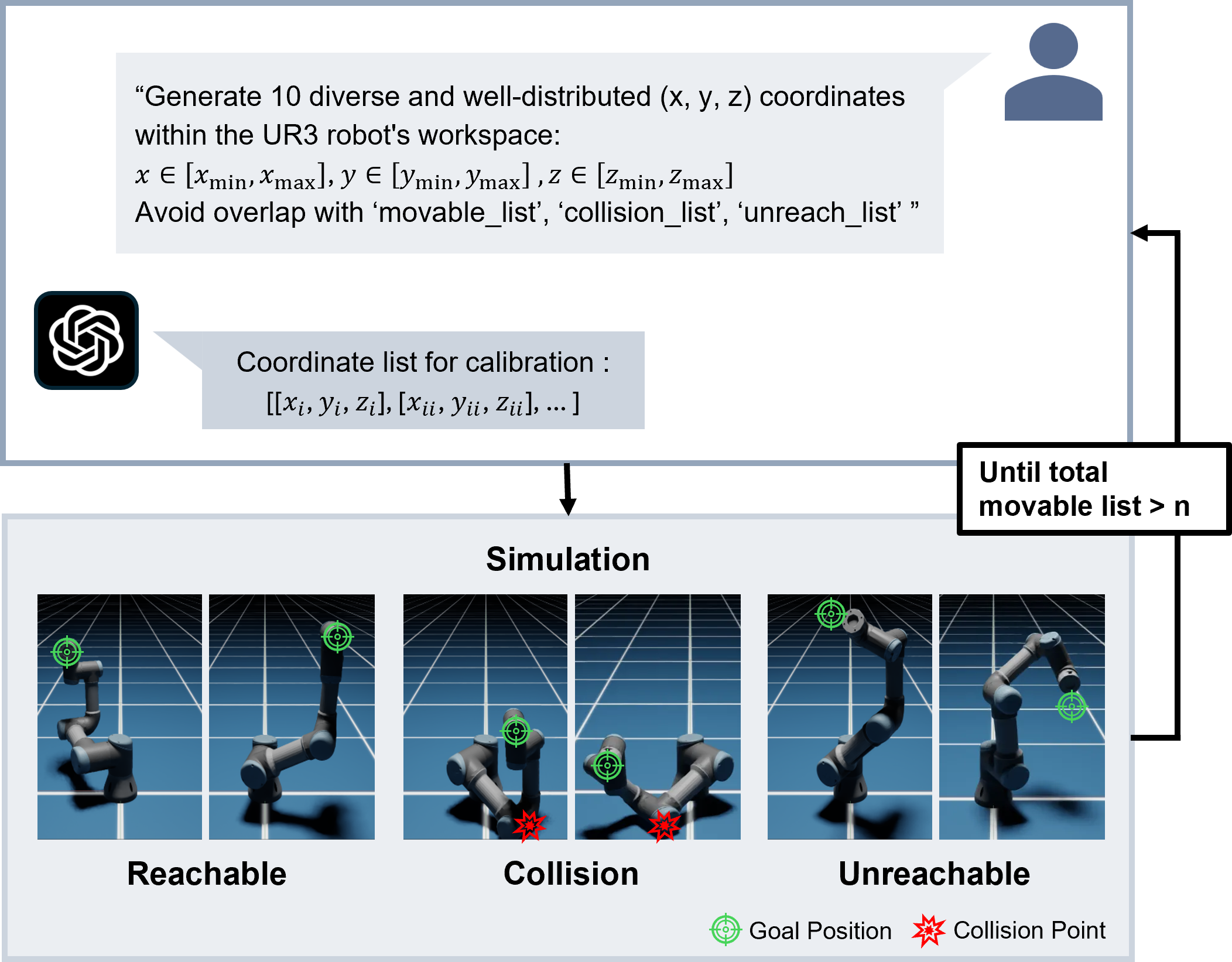}
    \caption{A pipeline where candidate positions are pre-validated in simulation for reachability and collision before being used in the closed-loop modular code generation process. Candidate positions are tested in simulation and classified as reachable, collision, or unreachable. These results are fed back to refine future generations until a sufficient number of reachable positions is obtained.}
    \label{fig:simul_gpt}
\end{figure}

\begin{equation*}
\mathcal{C}_{\text{gpt}} = G(\mathcal{W}, L_{\text{rea}}, L_{\text{col}}, L_{\text{unr}})
\end{equation*}

where $W \subset \mathbb{R}^3$ is the robot's workspace constraint, \\
$L_{\text{rea}}, L_{\text{col}}, L_{\text{unr}}$ are sets of previously observed reachable, collision, and unreachable positions, respectively, $G(\cdot)$ is a function that calls the LLM with prior outcomes as prompts, acting as a coordinate generator that produces candidate end-effector positions.

The candidate positions $C_{\text{gpt}}$ are then filtered through inverse kinematics (IK):
\[
C_{\text{gen}} = \{ c \in C_{\text{gpt}} \mid IK(c) \neq \emptyset \}
\]
where, $C_{\text{gen}}$ represents the subset of position candidates from $C_{\text{gpt}}$ that are kinematically reachable, i.e., those that pass IK filtering.  
Each $c \in C_{\text{gen}}$ denotes a single 3D end-effector position that satisfies this condition. IK serves both to reject infeasible samples and to provide executable joint trajectories.

The simulator evaluates each pose $c \in C_{\text{gen}}$ using a function $F_{\text{sim}}$, which classifies them into:
\[
(C_{\text{rea}}, C_{\text{col}}, C_{\text{unr}}) = F_{\text{sim}}(C_{\text{gen}})
\]
where: \\
$C_{\text{rea}}$: positions successfully reached without collisions \\
$C_{\text{col}}$: positions causing self-collision or collisions with the environment \\
$C_{\text{unr}}$: positions the robot fails to reach

These labeled results are re-fed into the LLM prompt to inform the next generation step, forming a feedback loop. The process repeats until enough reachable positions are collected. Through this process, the framework effectively integrates LLM-guided position generation with physics-based validation, enabling efficient collection of valid and executable robot coordinates.

Table~\ref{tab:coordinate} compares three approaches for calculating reachable coordinates in the robot workspace. The first is an analytical method using UR3 Denavit–Hartenberg (DH) parameters and joint limits. The second extends this by verifying inverse kinematics (IK) on the physical UR3 robot, yielding results closer to actual execution. The third is our simulation-based calibration method, where candidate coordinates are validated in simulation to assess reachability. 

Results show that the DH-based method has the lowest accuracy due to its reliance on analytical computation, while adding real-robot verification improves performance. Purely analytical approaches are prone to collision or infeasibility, whereas simulation-based validation provides the most reliable outcomes. 

In conclusion, the proposed framework integrates LLM-guided candidate generation with simulation-based validation, ensuring that only physically feasible and collision-free coordinates are retained. Simulation is employed exclusively in this validation stage, while the validated coordinates are subsequently used for real robot code generation, thereby enhancing execution fidelity.

\begin{table}[t!]
    \centering
    \caption{Comparison of reachable coordinate generation accuracy.}
    \begin{tabular}{c|c|c|c}
        \hline
        Metric & \makecell{Analytical\\(DH-based)} & \makecell{Analytical +\\Robot Verification} & Simulation-based \\ \hline
        Accuracy (\%) & 56.67 & 83.33 & 100.00 \\ \hline
    \end{tabular}
    \label{tab:coordinate}
\end{table}

\subsection{ModuLoop for Hand-eye Calibration}
\begin{figure*}
    \centering
    \includegraphics[width=0.95\linewidth]{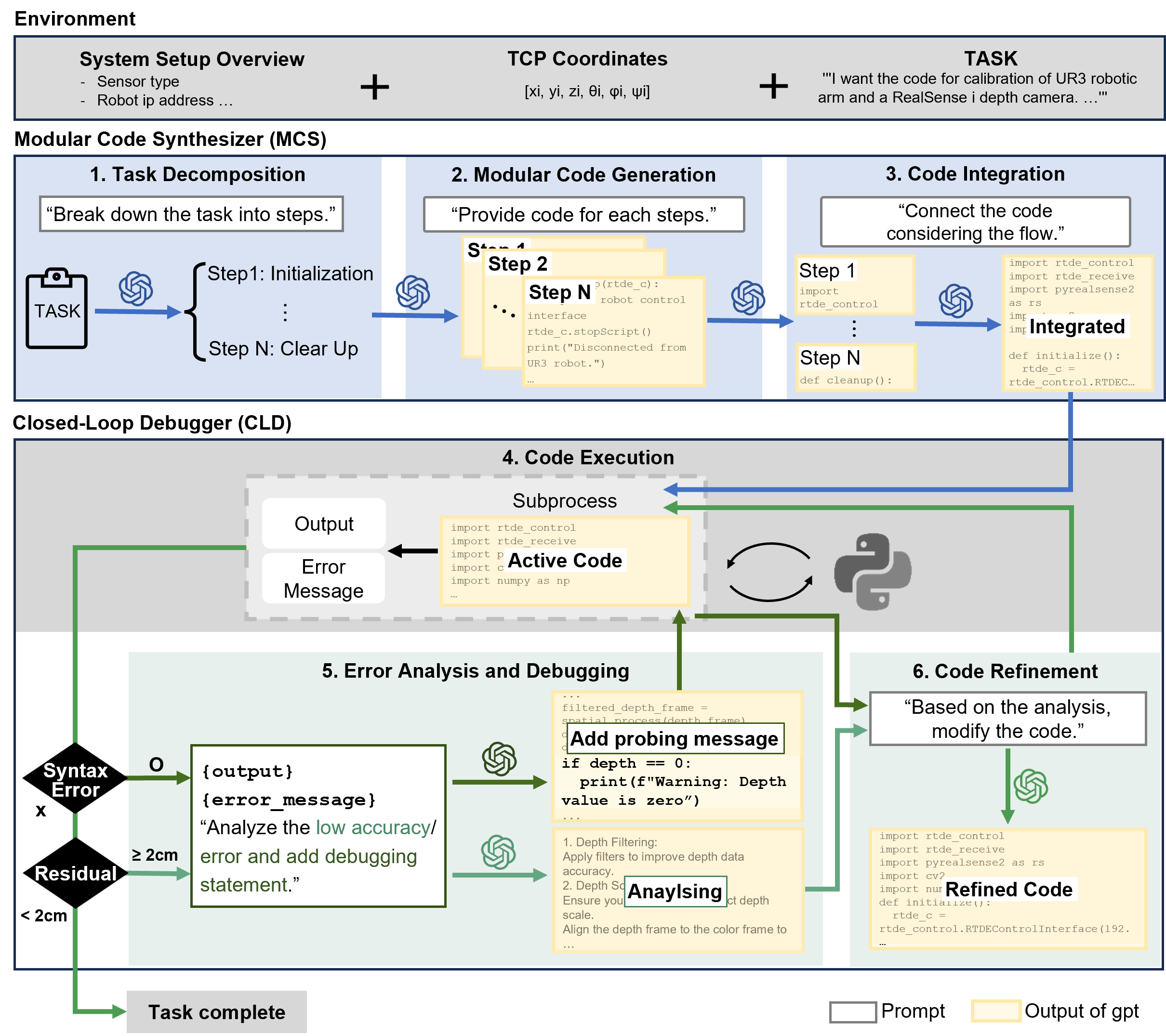}
    \caption{ModuLoop: A LLM-based control framework architecture for autonomously executing low-level robotic tasks from natural language instructions. Given a task, along with essential information such as reachable TCP coordinates (from simulation), sensor specifications, and robot IP, the framework proceeds through six stages. (1) The task is decomposed into sequential code-generation steps. (2) A code block is generated for each step. (3) The individual blocks are integrated into an initial executable script. (4) The code is executed via a subprocess for real-time interaction. (5) Errors are analyzed: syntax errors are handled through direct debugging, while low accuracy triggers debugging-message insertion and further diagnosis. (6) Based on analysis, the code is refined iteratively until both correctness and task accuracy criteria are met.}
    \label{fig:figure3}
\end{figure*}

A closed-loop framework is proposed to automate the calibration process between the robotic arm and the camera using GPT-4o. The methodology consists of two main stages: (1) Modular Code Synthesizer (MCS), where an initial Python script is automatically generated from a natural language prompt, and (2) Closed-Loop Debugger (CLD), in which the script is iteratively refined based on execution results. An overview of the entire workflow is shown in Fig.~\ref{fig:figure3}.

\subsubsection {Environment}
The process begins with the definition of the overarching goal of low-level control. In this case, the objective is to calculate the 3D-to-3D transformation matrix between the robot’s workspace and the camera’s coordinate system. To achieve this, a structured prompt is provided to GPT, containing essential information such as the robot’s IP address and sensor specifications. In addition, movable coordinates obtained from simulation are also included in the prompt.

\subsubsection{Modular Code Synthesizer}
The code generation stage systematically decomposes the calibration task and transforms it into automatically executable code. This process consists of three steps.

\begin{itemize}
    \item \textbf{Task Decomposition \& Planning:} The LLM decomposes the given calibration task into a series of independent subtasks. For example, to compute the transformation matrix, the procedure may include steps such as moving to predefined coordinates, acquiring sensor data, and performing matrix computation. Each subtask is then passed as a structured prompt to the next LLM instance.
    \item \textbf{Modular Code Generation:} Each subtask is implemented as a Python function or code block.
    \item \textbf{Code Integration:} The generated code blocks are assembled into a coherent executable script while maintaining logical consistency. This script is executed via subprocesses, enabling the system to observe robot motions and process camera data in real time.
\end{itemize}

\begin{algorithm}
\small
\caption{Closed-Loop Feedback for Calibration}
\KwIn{Initial code $C_0$; Evaluator $E$; Analyzer $A$; Refiner $R$; Max iterations $T$}
\KwOut{Refined calibration code $C^*$}

$C \leftarrow C_0$\;
$t \leftarrow 0$\;

\While{$t < T$}{
    /* \textit{Execute current code} */
    
    $($output $o$, error $e) \leftarrow \texttt{Run}(C)$ 
    
    \If{$e \neq \emptyset$}{
        /* \textit{If error occurs, insert probes and fix} */
        
        $C_d \leftarrow A.\texttt{add\_probing}(C, o, e)$

        $(o', e') \leftarrow \texttt{Run}(C_d)$

        $C \leftarrow R.\texttt{apply\_fix}(C_d, o', e')$
    }
    \Else{
        /* \textit{If no error, evaluate calibration quality} */
        
        result $r \leftarrow E.\texttt{evaluate}(o)$
        \If{r == ``inaccurate''}{
            /* \textit{If inaccurate, diagnose and refine} */
            $diag \leftarrow A.\texttt{analyze\_accuracy}(C, o)$
            $C \leftarrow R.\texttt{apply\_improvement}(C, o, diag)$
        }
        \ElseIf{r == ``calibration success''}{
            /* \textit{If successful, return calibrated code} */
            
            \Return{$C$}
        }
    }

    $t \leftarrow t + 1$\;
}
\Return{Failure: calibration did not converge}
\label{alg:closed-loop}
\end{algorithm}

\begin{figure}[h]
    \centering
    \includegraphics[width=1.0\linewidth]{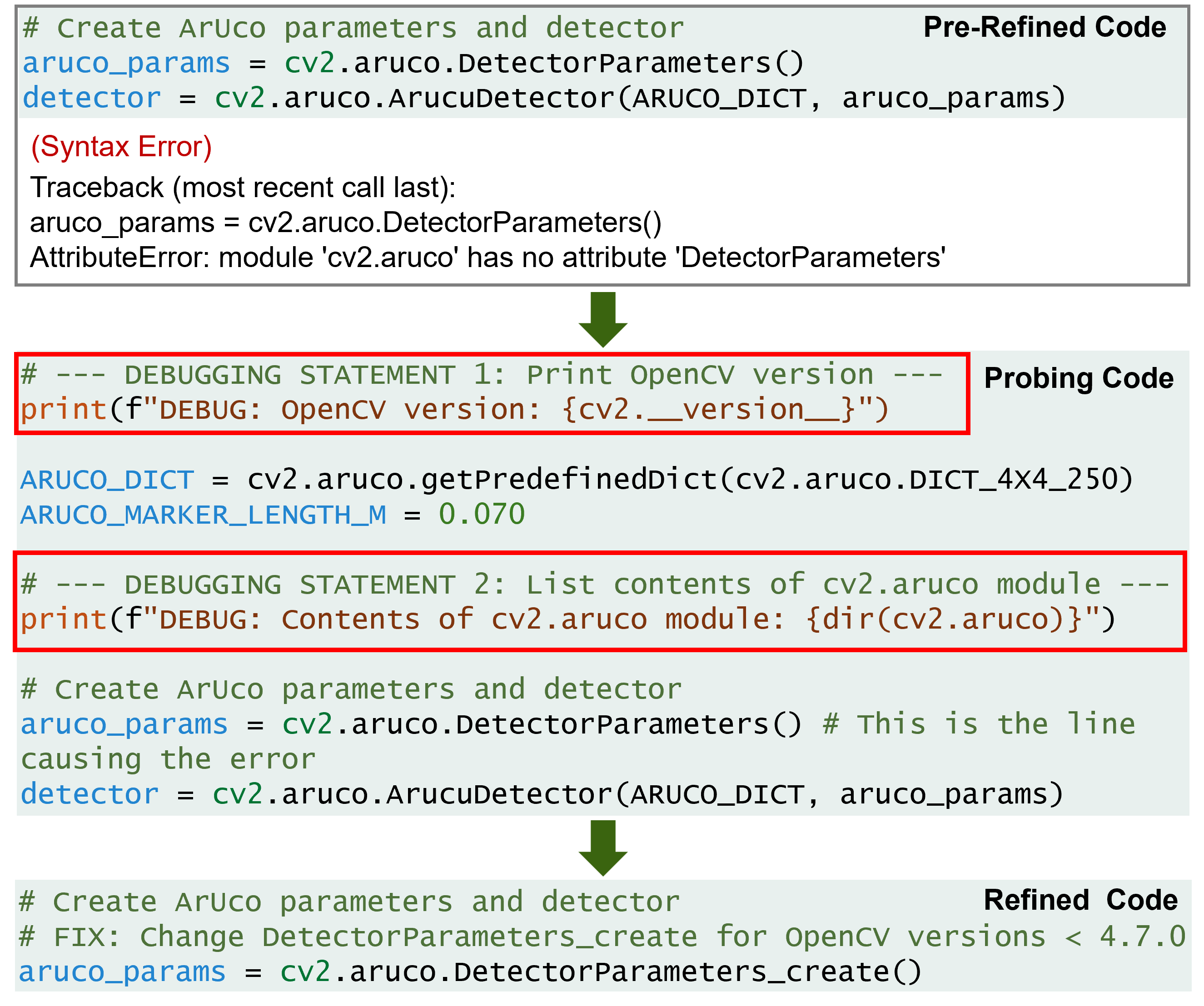}
    \caption{Syntax Error Debugging in Closed-Loop Process. This case illustrates the process of resolving a version mismatch error in OpenCV’s ArUco module. Based on the pre-refined code and its output message, probing code is generated to identify the root cause, and the issue is resolved by refining the code accordingly.}
    \label{fig:modul_result1}
\end{figure}

\begin{figure}
    \centering
    \includegraphics[width=1.0\linewidth]{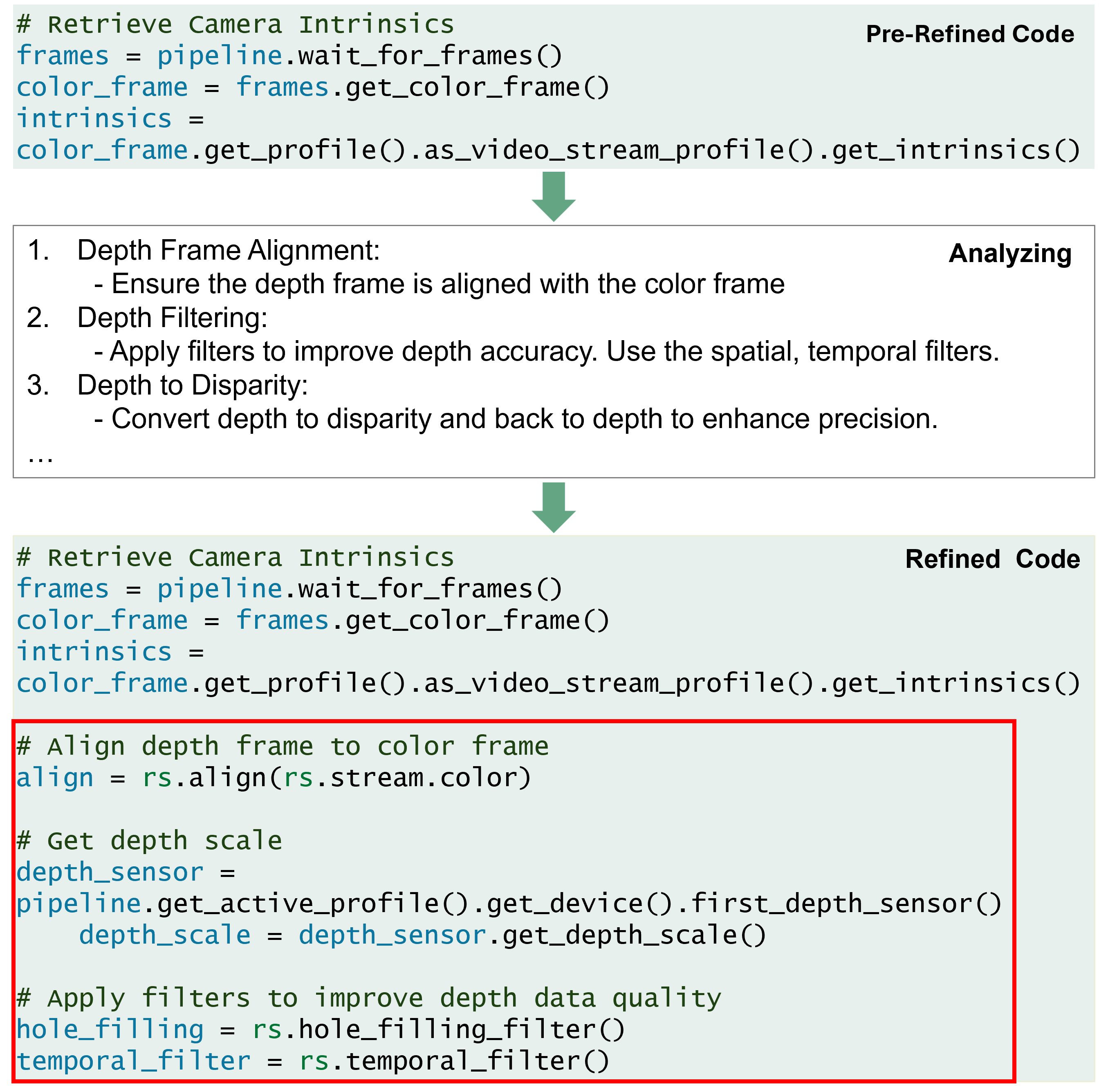}
    \caption{Accuracy Improvement Debugging in Closed-Loop Process. This case demonstrates how adding depth alignment and filtering to the camera pipeline improves analysis accuracy. Based on the initial code and its output messages, the system analyzes potential causes of low accuracy and recommends refinement strategies.}
    \label{fig:modul_result2}
\end{figure}

\subsubsection{Closed-Loop Debugger}
The code generation stage provides an initial Python script, but in real robotic environments, unexpected runtime errors or insufficient accuracy may prevent reliable execution. To address these issues, we propose a closed-loop feedback structure that diagnoses execution outcomes and iteratively refines the code (Algorithm~\ref{alg:closed-loop}). In this process, the LLM acts not only as a code generator but also as an active participant in a continuous loop of execution, analysis, and refinement.

At each iteration, the current code \( C \) is executed in the robotic environment, producing an output \( o \) and an error message \( e \). If an error occurs, the Analyzer formulates hypotheses about possible causes and re-executes the code with probing messages inserted to validate these hypotheses. The results are then passed to the Refiner, which applies targeted modifications accordingly. This process is exemplified in Fig.~\ref{fig:modul_result1}, which shows a syntax error case in OpenCV’s ArUco module resolved through probing and refinement.

If no runtime error is detected, the Evaluator assesses calibration accuracy based on the output messages. The output includes either “calibration success” or “calibration inaccurate.” If classified as “inaccurate,” the Analyzer diagnoses potential causes, and the Refiner updates the code to improve accuracy. An example of this process is shown in Fig.~\ref{fig:modul_result2}, where calibration accuracy is improved through additional alignment and filtering strategies.

This loop continues until one of two termination conditions is met: (1) the calibration reaches the required accuracy, in which case the refined script \( C^* \) is returned, or (2) the maximum number of iterations \( T \), is reached, in which case the process is considered a failure. Through this mechanism, the LLM functions as a self-correcting agent, capable of iterative adaptation, real-time diagnosis, and execution-driven refinement throughout the calibration process.

\section{Performance OF Modular Code generation and Closed-Loop Debugger}

\begin{table*}[h]
\centering
\caption{Comparison of calibration performance across different code generation and feedback strategies. The iteration was terminated and deemed unsuccessful after ten attempts. Success Cases: calibration error $<$ 0.02 m. 
Abbreviations: CaP - Code as Policies, SPG – Single Path Generator, MCS – Modular Code Synthesizer.}
\begin{tabular}{l|l|c|c|c|c}
\hline
\multicolumn{2}{c|}{\textbf{Model}} & 
%\multicolumn{1}{c|}{\textbf{Backbone}} &
\begin{tabular}[c]{@{}c@{}}Code Gen\\ Success (\%)\end{tabular} & 
\begin{tabular}[c]{@{}c@{}}Accuracy\\ Success (\%)\end{tabular} & 
\begin{tabular}[c]{@{}c@{}}Verified Error (Success Cases, m)\\ mean / median\end{tabular} &
\begin{tabular}[c]{@{}c@{}}Verified Error (All Cases, m)\\ mean / median\end{tabular} \\
\hline
& CaP with Basic Robot Control API & - & - & - & - \\
&CaP with Full API & 90.0 & - & - & 0.038/ 0.039 \\
&ProgPrompt& 66.67 & - & - & 0.026/ 0.026 \\
&SPG & - & - & - & - \\
GPT-4o & SPG + Debugger & 6.67 & - & - & 0.802/ 0.802  \\
& SPG + Debugger w/ Probing &10.0 & - & - & 0.485/ 0.062 \\
& MCS & 23.33 & 6.67 & 0.029/ 0.029 & 0.308/ 0.283 \\
& MCS + Debugger  & 73.33 & 60.0 & 0.174/ 0.019  & 0.074/ 0.014\\
& MCS + Debugger w/ Probing  & 96.67 & 86.67 & 0.048/ 0.011 & 0.170/ 0.012 \\
\hline
GPT-4.1 mini & MCS + Debugger w/ Probing & 73.33  & 40.0 & 0.085/ 0.068 &  0.337 /0.068\\
Gemini & MCS + Debugger w/ Probing & 90.0 & 73.33 & 0.064/ 0.064  & 0.091/ 0.065 \\
\hline
\end{tabular}
\label{tab:result_table_v2}
\end{table*}
\subsection{Experimental Setup}
This experiment was designed to quantitatively evaluate the performance of two key components of the proposed closed-loop calibration framework: MCS and CLD. The objective is to assess their contribution to the robustness of code execution and calibration accuracy.

The experiment was conducted using a UR3 robotic manipulator and an Intel RealSense D435i depth camera. An ArUco marker was attached to the robot’s tool center point (TCP), and the camera was positioned 1.3 meters in front of the robot, facing it directly. The robot is controlled through the RTDE (Real-Time Data Exchange) interface, and the control system operates at a frequency of 125 Hz. Communication between the robot and the control system is conducted via TCP/IP over a local network.

To assess the effectiveness of the MCS and CLD, we included an additional model to provide a baseline for comparison. This model, referred to as the Single-Path Generator (SPG), generates the entire code in a single pass without any task planning, directly producing code from the given task description in a single prompt. In the feedback stage, two additional comparison models were included.
The first is an open-loop model that performs no feedback at all,
and the second is a limited closed-loop model without probing.
The latter follows the closed-loop structure but does not insert diagnostic code
to observe runtime behavior, relying solely on the final execution results as feedback.
Thus, it represents a minimally applied feedback loop. All experiments were repeated 30 times to ensure statistical reliability.

\begin{comment}
The experimental results are summarized in Table~\ref{tab:result_table_v2}. As shown, the model combining MCS and the CLD achieved the highest success rate and accuracy. In contrast, the single-prompt approach exhibited substantially lower performance, suggesting that generating code without task decomposition can lead to structurally and logically incomplete programs.

Fig.~\ref{fig:loop_boxplot} illustrates the effect of integrating the CLD with the MCS. Through probing-based debugging, a larger number of initial samples were successfully calibrated, and the average number of iterations required for convergence was reduced. This indicates that error-hypothesis probing improves calibration efficiency. In summary, MCS enables structured code generation, while the Debugger iteratively refines it to maximize overall performance.
\end{comment}

\subsection{Experimental Results}
Each model was evaluated according to the following three criteria:

\begin{enumerate}
    \item Successful code generation of without syntax errors
    \item Satisfaction of a predefined accuracy threshold ($< 2$ cm)
    \item Positional error between the robot’s actual TCP position and the transformed position computed from the camera
\end{enumerate}

The experimental results are summarized in Table~\ref{tab:result_table_v2}. The combination of the Modular Code Synthesizer (MCS) and the Closed-Loop Debugger (CLD) achieved the highest success rate and accuracy, whereas the single-prompt approach showed substantially lower performance, indicating that task decomposition is critical for generating structurally and logically complete programs.

\begin{figure}
    \includegraphics[width=1\linewidth]{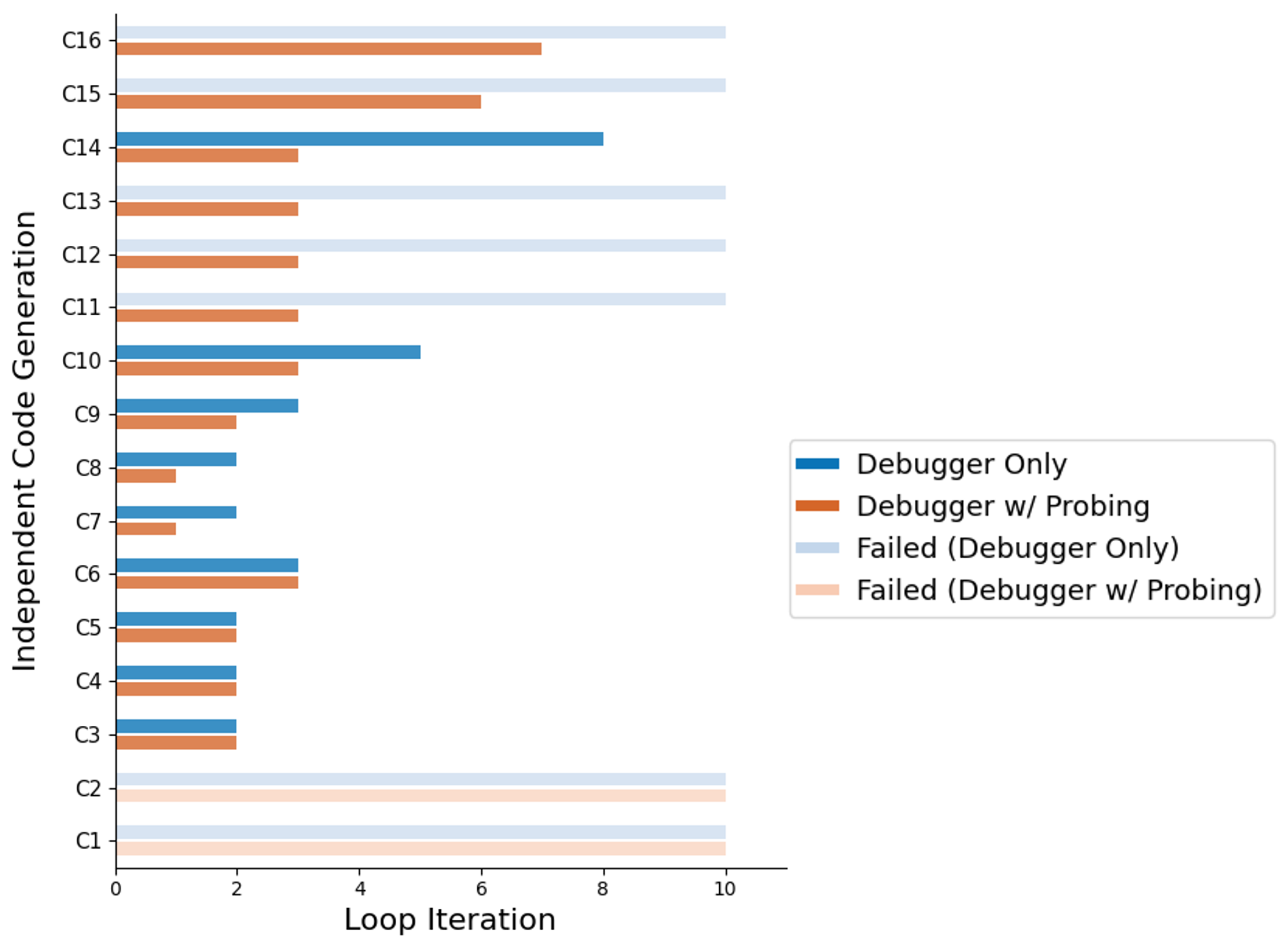}
    \caption{Comparison of feedback loop iterations between models with and without the proposed Closed-Loop Debugger. Each bar represents the number of refinement iterations required for successful task completion across 16 code generation experiments.}
    \label{fig:loop_boxplot}
\end{figure}

As shown in Fig.~\ref{fig:loop_boxplot}, integrating the CLD with the MCS improved calibration efficiency: Through probing-based debugging, more initial samples were successfully calibrated, and fewer iterations were required for convergence. This demonstrates that error-hypothesis probing accelerates convergence and enhances robustness

This study compared ModuLoop with ProgPrompt~\cite{singh2023progprompt} and Code as Policies (CaP)~\cite{liang2023code} on the same hand–eye calibration task. ProgPrompt is a prompting technique that leverages code structures instead of natural language, where available robot actions and environmental objects are presented as Python functions and lists. By providing example tasks together with these definitions, the LLM is guided to generate new task plans directly in the form of code.

Code as Policies is a paradigm where the LLM constructs control code around user-defined APIs, generating programs mainly by invoking provided APIs and, when necessary, defining new ones recursively. In our experiments, we evaluated two conditions. In \textit{CaP with Full API}, the LLM was given a rich set of APIs covering not only low-level robot and sensor operations but also calibration procedures, along with hints. In contrast, \textit{CaP with Basic Robot Control API} provided only simple functions for primitive robot control, while essential information as in ModuLoop was supplied for the rest.

ProgPrompt and CaP with Full API achieved relatively strong performance but were constrained by their reliance on predefined APIs and the absence of feedback for error correction. ModuLoop, by comparison, does not depend on predefined APIs and directly generates low-level control code, which is iteratively refined through feedback, thereby demonstrating robustness and intuitive applicability even in precise robotic tasks.

We further evaluated ModuLoop with different LLM variants. GPT-4.1-mini achieved slightly lower success rates than GPT-4o, but demonstrated advantages in cost and response speed. This indicates that it can serve as a cost- and time-efficient alternative, albeit with reduced reliability. Gemini achieved code-generation and accuracy comparable to GPT-4o, but the derived transformation matrices fell below the required accuracy threshold, indicating limited calibration stability.

\section{Evaluation of the generalizability and practicality of the closed-loop code generation and debugging framework}

To validate the calibration-derived transformation matrix and assess ModuLoop’s generality, we applied it to a representative pick-and-place task involving object recognition, coordinate alignment, motion planning, and grasp planning—all handled through LLM-based reasoning and code generation. The LLM received natural language instructions, target poses from perception, and motion planning guidelines.

Unlike calibration, where closed-loop debugging was feasible, executing pick-and-place trials risked collisions and environmental changes. We therefore adopted static debugging, supplying GPT-4o with a structured evaluation checklist to anticipate and reason about potential failure cases.

\setlength{\parskip}{0pt} 
\subsection{Object detection for pick and place}

When a natural language instruction is given, the system executes the task through a multi-stage pipeline. RGB-D images from front and gripper-mounted cameras are processed by the Segment Anything Model~\cite{kirillov2023segment}, which segments object masks and extracts their center coordinates.

\begin{figure}[h]
    \centering
    \includegraphics[width=1.0\linewidth]{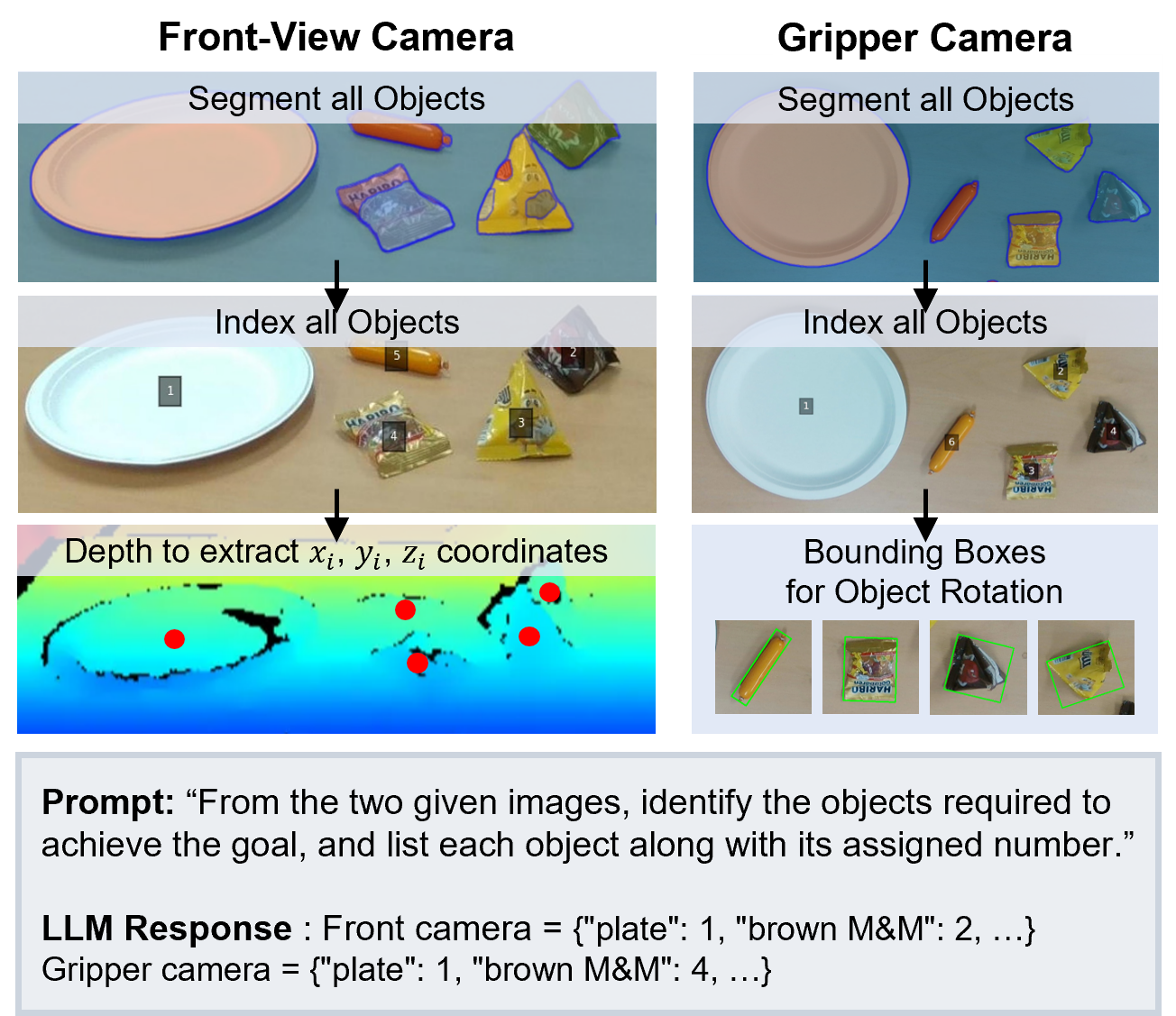}
    \caption{LLM-based Object Recognition for Pick-and-Place Tasks. RGB-D images from both front and gripper-mounted cameras are processed by the Segment Anything Model (SAM) to segment object masks and extract their center coordinates. The annotated images and the natural-language command are then provided to the LLM, which contextually interprets the scene to identify the target object and determine whether rotation is required for precise manipulation.}
    \label{fig:object_detection}
\end{figure}

Each object is assigned a unique label, and the annotated images with the user command are provided to the LLM, which contextually interprets the scene to identify target objects rather than performing simple classification. For example, given “I’m hungry. Can you get me some snacks?”, the LLM selects relevant items such as “sausages” or “chocolate,” returning their names and IDs from each view (Fig.~\ref{fig:object_detection}).

The LLM also considers object geometry and context to decide if rotation is needed for precise pick-and-place execution.

\subsection{Quantitative Evaluation of ModuLoop on Pick-and-Place Tasks}
\begin{comment}
We evaluated ModuLoop on five pick-and-place tasks of increasing complexity and linguistic difficulty (see Table \ref{tab:Task Requirements}), each executed 25 times using three code generation methods:
\begin{itemize} \item \textbf{Method 1}: Single-Prompt Code Generation
\item \textbf{Method 2}: Modular Code Synthesizer
\item \textbf{Method 3}: ModuLoop
\end{itemize}

\end{comment}

We evaluated ModuLoop on five pick-and-place tasks of increasing complexity and linguistic difficulty (see Table~\ref{tab:Task Requirements}), each executed 25 times using three code generation methods: (1) Single-Prompt Code Generation, (2) Modular Code Synthesizer, and (3) ModuLoop.

\begin{figure*}
    \centering
    \includegraphics[width=1.0\linewidth]{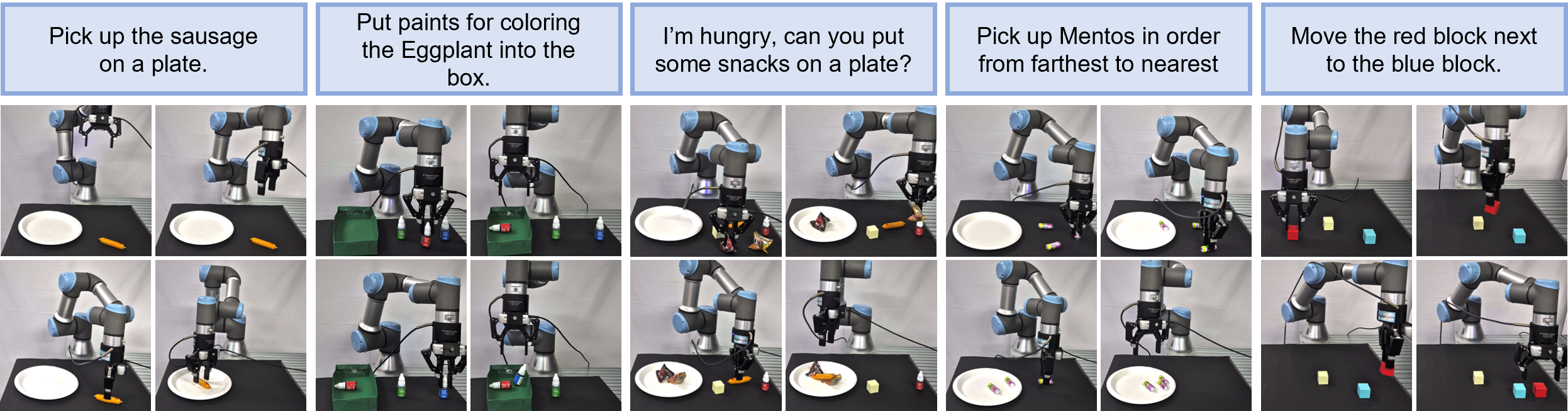}
    \caption{Pick-and-place tasks performed by a robot based on language instructions. Each task is visualized as a 2×2 image sequence, ordered from top-left to bottom-right to show the temporal progression of execution.}
    \label{fig:pnp}
\end{figure*}

The tasks were specifically structured to assess the system's capabilities in perception, reasoning, and control, with increasing complexity. in Fig.~\ref{fig:pnp}. shows example scenes of a robotic arm performing each of the five pick-and-place tasks based on natural language instructions. The main characteristics of each task are as follows:

\begin{itemize} \item \textbf{Simple 1}: Basic object targeting and orientation control
\item \textbf{Moderate 1}: Logical reasoning involving multiple objects and inference of color mixing
\item \textbf{Moderate 2}: Semantic categorization of objects and orientation control
\item \textbf{Moderate 3}: Distance-based ordering and applying appropriate orientations for identical objects located at different positions
\item \textbf{Hard 1}: Spatial reasoning, precise relative positioning, and orientation control
\end{itemize}

As shown in Table~\ref{tab:Pnp_success_rate}, ModuLoop achieved the highest success rates, particularly in complex tasks (Tasks 3–5) requiring semantic reasoning, spatial understanding, and precise motion planning, outperforming the Single-Prompt baseline. These results underscore the value of structured command decomposition and GPT-based debugging for translating complex manipulation commands into executable code.

While this study confirms the feasibility of LLM-based low-level code generation in calibration and pick-and-place tasks, we acknowledge a key limitation: ModuLoop currently relies only on minimal APIs that provide object positions, limiting its applicability to relatively simple tasks. Contact-rich manipulation, such as assembly or opening a drawer, requires richer environmental understanding—including perception modules, motion planners, and value map composition provided as APIs as exemplified in works such as VoxPoser~\cite{huang2023voxposer}.

\begin{table*}[]
    \centering
    \caption{Capabilities required by each task to guide LLM-based code generation.}
    \begin{tabular}{c|c|c|c|c|c|c|c}
        \hline
        \multirow{3}{*}{Task ID} & \multirow{3}{*}{Task Description} & \multicolumn{6}{c}{Key Capabilities Required per Task (increasing complexity)} \\ \cline{3-8}
        & & Class & Logical & Multi & Orientation & \multirow{2}{*}{Sequencing} & Spatial \\
        & & Reasoning & Inference & Object & Control & & Reasoning \\ \hline
        \multicolumn{1}{l|}{Simple 1} & \multicolumn{1}{l|}{Pick up the sausage on a plate.} & & & &\checkmark & & \\ 
        \multicolumn{1}{l|}{Moderate 1} & \multicolumn{1}{l|}{Put paints for coloring the Eggplant into the box.} & &\checkmark &\checkmark & & & \\ 
        \multicolumn{1}{l|}{Moderate 2} & \multicolumn{1}{l|}{I’m hungry, can you put some snacks on a plate?} &\checkmark & &\checkmark &\checkmark & & \\ 
        \multicolumn{1}{l|}{Moderate 3} & \multicolumn{1}{l|}{Pick up Mentos in order from farthest to nearest.} & & &\checkmark &\checkmark &\checkmark & \\ 
        \multicolumn{1}{l|}{Hard 1} & \multicolumn{1}{l|}{Move the red block next to the blue block.} & & &\checkmark &\checkmark & &\checkmark \\ \hline
    \end{tabular}
    \label{tab:Task Requirements}
\end{table*}

\begin{table}[]
    \centering
    \caption{Success Rates(\%) of Code Generation Methods Across Tasks}
    \begin{tabular}{c|c|c|c}
        \hline
         \multirow{2}{*}{Task ID}&  Single & Modular Code& \multirow{2}{*}{ModuLoop}\\
         &Code Generator&Synthesizer&\\ \hline
         \multicolumn{1}{l|}{Simple 1}&  60& 88& \textbf{96}\\
         \multicolumn{1}{l|}{Moderate 1}&  64& 80& \textbf{96}\\
         \multicolumn{1}{l|}{Moderate 2}&  36& 60& \textbf{92}\\
         \multicolumn{1}{l|}{Moderate 3}&  12& 76& \textbf{92}\\
         \multicolumn{1}{l|}{Hard 1}&  24& 48& \textbf{60}\\ \hline
    \end{tabular}
    \label{tab:Pnp_success_rate}
\end{table}

\section{Conclusion}
This study verified that pre-trained large language models (LLMs) can autonomously generate and execute low-level robot control code without further training. The ModuLoop pipeline demonstrated high execution accuracy in camera-to-robot calibration and pick-and-place tasks, confirming the feasibility of LLM-based control in physical environments.

By positioning the LLM as the central agent in the control process, this work demonstrates the feasibility of using pre-trained LLMs to translate natural language into robot-executable code. Its ability to handle complex control logic without manual programming or task-specific training underscores the approach’s practicality. However, due to the inherent structure of LLMs, generating complete control code introduces noticeable latency, which can limit responsiveness.

In future work, we will extend the framework to real-world industrial applications such as process and factory automation, and evaluate its generality and hardware-independence across diverse robotic platforms beyond the UR3.

\begin{comment}
This study verified whether pre-trained large language models, without additional training, can autonomously generate and execute low-level robot control code. We proposed the ModuLoop pipeline and validated it on two representative tasks: camera-to-robot calibration and pick-and-place manipulation, achieving high execution accuracy and demonstrating the feasibility of LLM-based control in physical environments.

By positioning the LLM as the central agent, ModuLoop shows that natural language can be directly translated into robot-executable code, handling complex control logic without manual programming or task-specific training. A current limitation is the non-negligible latency in generating complete code.

Future work will extend ModuLoop to industrial automation scenarios and evaluate its generality and hardware independence across diverse robotic platforms.

\end{comment}

\bibliographystyle{IEEEtran}
\bibliography{reference}

\newpage
\vfill
\end{document}